\documentclass[lettersize, 10pt, journal, twoside]{IEEEtran}
\usepackage[T1]{fontenc}

\usepackage{amsmath,amssymb,amsfonts}
\usepackage{graphicx}

\usepackage{amsthm}
\usepackage{xcolor}
\usepackage[figuresright]{rotating}

\usepackage{graphicx}
\graphicspath{{/}{fig/}}

\usepackage{array}
\usepackage{textcomp}
\usepackage{xcolor}
\usepackage{multirow}
\usepackage{booktabs}

\usepackage{mathtools}
\usepackage{breqn}
\usepackage{float}
\usepackage{pifont}
\usepackage{filecontents}

\usepackage{pgfplots}
\pgfplotsset{compat=1.18}
\usepgfplotslibrary{groupplots}
\usepackage{svg}

\usepackage{caption}
\usepackage{orcidlink}

\usepackage{multirow,tabularx}
\usepackage{hyperref}
\usepackage{flushend}
\usepackage{algorithmic}
\usepackage[vlined, ruled, shortend]{algorithm2e}

\definecolor{AviaColor}{RGB}{228,26,28}
\definecolor{Mid360Color}{RGB}{55,126,184}
\definecolor{OusterColor}{RGB}{77,175,74}

\usepackage{color,soul}

\usepackage{xcolor}
\newif\ifrevision
\revisiontrue   

\ifrevision
  
\else
  
\fi

\usepackage{balance}

\usepackage{enumitem}

\newlength\figureheight
\newlength\figurewidth
\SetAlCapNameFnt{\footnotesize}
\SetAlCapFnt{\footnotesize}

\title{
    Offline Vision-Language Navigation with Geometric Goal Localization for Outdoor Environments

\author{
        Ali Salmasi\,\orcidlink{0009-0001-3982-9962},
        Xianjia Yu\,\orcidlink{0000-0002-9042-3730},
        Tomi Westerlund\,\orcidlink{0000-0002-1793-2694}
    \thanks{This research is supported by the European Union’s Horizon Europe research and innovation programme under the Marie Skłodowska-Curie Actions grant agreement No. 101125250.}
    \thanks{
    All authors are with \href{https://tiers.utu.fi}{Turku Intelligent Embedded and Robotic Systems (TIERS) Lab}, University of Turku, Turku, Finland.(e-mail:\{ali.a.salmasi, xianjia.yu, tovewe\}@utu.fi.)}
}
}

\begin{document}

\maketitle



\begin{abstract}\label{sec:abstract}%
Foundation-model-based vision-language navigation (VLN) has substantially advanced autonomous robot navigation by enabling robots to interpret natural-language instructions, identify semantic goals, and follow user-specified behavioral rules. However, existing VLN systems rely heavily on cloud-hosted foundation models for language understanding and semantic grounding, limiting their applicability where network connectivity is unavailable and reliable metric goal localization is required. Although recent small language models (SLMs) enable fully onboard inference, their suitability for robotic navigation, particularly for navigation instruction understanding and decomposition, has not been systematically evaluated. Meanwhile, achieving reliable semantic-to-metric goal localization without cloud services remains a practical challenge for outdoor VLN.

To address these challenges, this paper makes three contributions toward fully onboard VLN for outdoor environments.
First, we present the first systematic benchmark of 17 edge-deployable SLMs against 4 online APIs for robotic navigation instruction decomposition, evaluating their accuracy and latency on human annotated natural-language instructions across three computing platforms, providing practical guidance for selecting onboard language models. 
Second, we propose a lightweight hybrid semantic-geometric goal localization framework that combines open-vocabulary object detection, prompted segmentation, and LiDAR geometry to estimate metric goals while maintaining visual bearing guidance when reliable geometric observations are unavailable.
Finally, we integrated these advances into Edge-BehAV, a fully onboard extension of the BehAV architecture that enables cloud-independent behavior-guided navigation through onboard language understanding, perception, and hybrid semantic-geometric goal localization. Experimental results show that the best offline SLM matches the instruction decomposition performance to the strongest cloud API while running approximately $9\times$ faster without network connectivity. The proposed geometric goal localization framework reduces mean goal-distance error from 2.05\,m to 0.20\,m with lower computational cost. The complete system successfully performs 31/32 closed-loop outdoor navigation trials.

\end{abstract}

\IEEEpeerreviewmaketitle


\section{Introduction}\label{sec:introduction}
\IEEEPARstart{A}{utonomous} mobile robots are increasingly expected to operate in complex outdoor environments while interpreting high-level human intent expressed through natural-language instructions and adhering to context-sensitive behavioral rules~\cite{shah2023lmnav,10802716,weerakoon2025behav}.
Recent advances in foundation models, particularly large language models (LLMs) and Vision-Language Models (VLMs) have significantly advanced vision-language navigation (VLN) by enabling robots to interpret natural-language instructions, identify semantic landmarks, and comply with user-specified behavioral rules~\cite{8578485,gu-etal-2022-vision,ahn2022saycan}. These capabilities have broadened the applicability of mobile robots to environmental monitoring, infrastructure inspection, precision agriculture, and search-and-rescue missions, where users naturally specify tasks such as \textit{''navigate to the orange cone while avoiding pedestrians and staying on sidewalks''} rather than explicit metric coordinates.

\begin{figure}[t]
    \centering
    \includegraphics[width=0.425\textwidth]{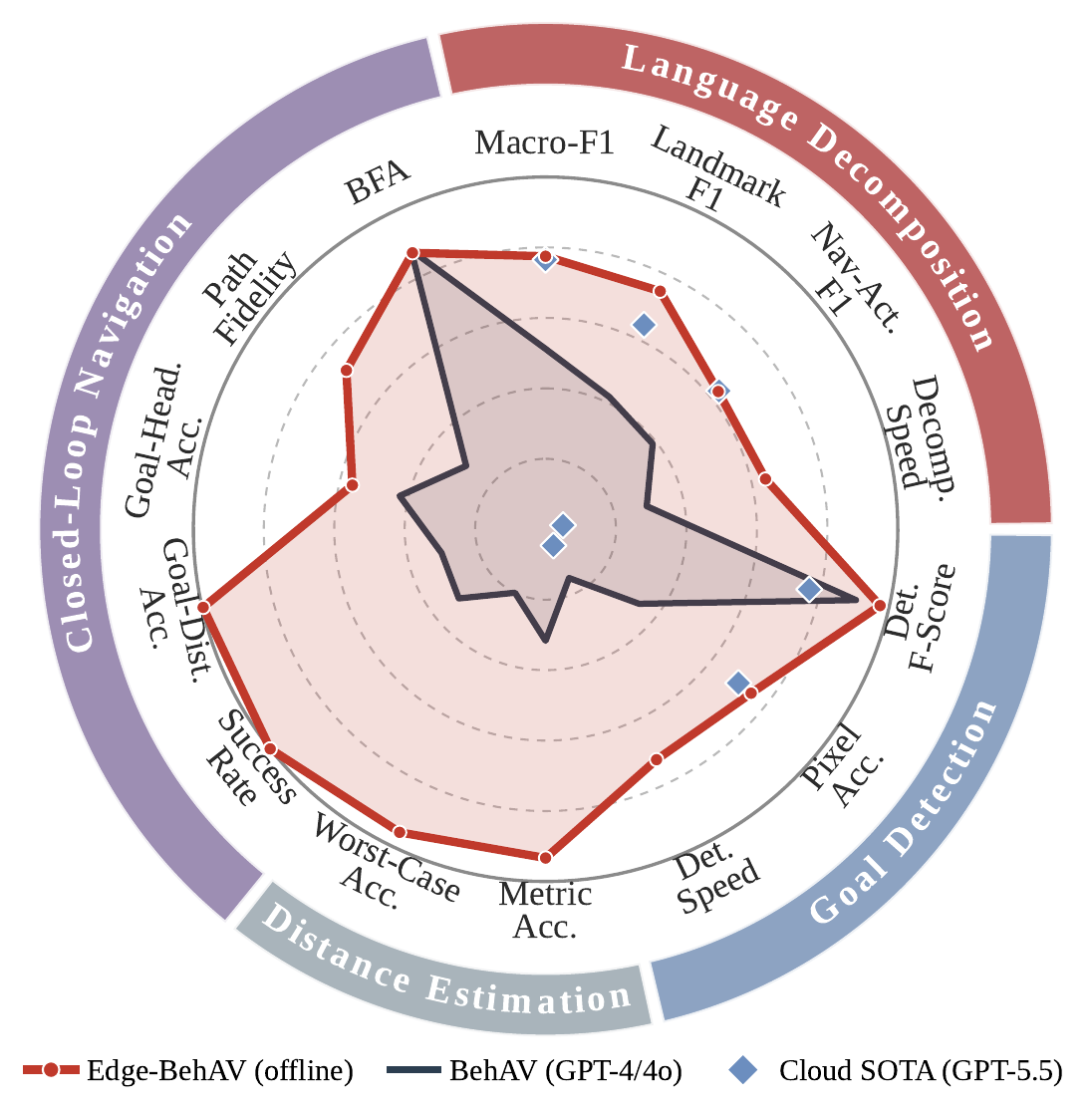}
    \caption{Edge-BehAV at a glance. Across all four evaluated capabilities, the fully offline system (red) matches or exceeds the original online BehAV pipeline (dark, GPT-4/GPT4o) and the strongest current cloud API (GPT-5.5, blue), while running entirely on-board. Axes are normalised so the outer ring is ideal.}
    \label{fig:demo}
    \vspace{-2.0em}
\end{figure}

Despite these advances, deploying vision-language navigation in real outdoor environments remains challenging. Existing foundation-model-enabled navigation systems remain heavily dependent on cloud-hosted language and vision-language models for instruction understanding and semantic grounding. This reliance may introduce high latency, requires continuous network connectivity, and prevents reliable deployment in connectivity-constrained outdoor environments. 
Moreover, many existing systems estimate semantic goal locations through heuristic visual reasoning rather than sensor-grounded geometric localization, limiting metric accuracy and navigation reliability. BehAV~\cite{weerakoon2025behav}, for example, combines online GPT-4 for language decomposition, GPT-4o for visual landmark estimation, and CLIPSeg for real-time behavioral cost-map generation, demonstrating the effectiveness of foundation models for outdoor navigation. However, its reliance on cloud-hosted foundation models introduces multi-second inference latency (3--5\,s per VLM query in our experiments shown in Section~\ref{sec:results}), while its heuristic visual goal estimation illustrates the broader challenge of accurate semantic-to-metric goal localization for fully onboard outdoor VLN.

Recent advances in edge-deployable small language models (SLMs)~\cite{lu2024slmsurvey,abdin2024phi3,qwen2025report} and lightweight VLMs~\cite{chu2023mobilevlm,xiao2024florence2} have enabled competitive onboard reasoning on embedded robotic platforms. These developments create new opportunities for fully onboard vision-language navigation, but their practical deployment raises important questions.
To the best of our knowledge, no prior work has systematically benchmarked edge-deployable offline foundation models for robotic navigation instruction understanding across diverse model families, embedded platforms, and realistic navigation tasks. Such a benchmark is important because selecting foundation models for embedded robotic platforms requires understanding the trade-offs between instruction understanding quality, inference latency, and computational resources rather than relying solely on general-purpose language benchmarks.
At the same time, efficient onboard semantic grounding and accurate geometric goal localization remain practical challenges for fully offline vision-language navigation.


To address these limitations, this paper develops a fully onboard VLN approach for outdoor environments by fusing systematic evaluation of edge-deployable language models, hybrid semantic-geometric goal localization, and their integration into a complete navigation system.
To validate the proposed approach, we implement it as Edge-BehAV, a fully onboard extension of the BehAV architecture that replaces cloud-dependent language understanding and perception with efficient onboard alternatives while introducing sensor-grounded geometric goal localization.
Fig~\ref{fig:demo} summarizes the capabilities of the proposed system relative to cloud-based alternatives.
The contributions of this work are as follows:
\begin{enumerate}[label=\roman*)., leftmargin=*, itemindent=2em, labelsep=0pt]
\item \textbf{Navigation-specific Benchmark of Edge-Deployable Offline SLMs:} 
We present a navigation-specific benchmark of edge-deployable foundation models by evaluating 17 local and 4 cloud-API models across 350 human-annotated navigation instructions and three computing platforms.
The annotated instruction set will be released as an open-source benchmark to facilitate future validation of newly emerging models, providing practical guidance for selecting onboard language models for robotic navigation.
\item \textbf{Hybrid Semantic-Geometric Goal Localization:} We develop a hybrid semantic-geometric goal localization framework that combines open-vocabulary detection, prompted segmentation, and LiDAR geometry to estimate metric goals reducing the mean goal localization error from 2.05 m to 0.20 m. When reliable LiDAR observations are unavailable, the system maintains visual bearing guidance until sufficient geometric information becomes available.
\item \textbf{Fully Onboard System Integration and Validation:}  We implemented the proposed framework as Edge-BehAV by extending the BehAV architecture and validate it through 32 real-world closed-loop outdoor navigation trials on a Clearpath Husky A200, demonstrating that reliable behavior-guided navigation can be achieved entirely onboard without cloud-hosted foundation models.
\end{enumerate}

\section{Related Work} \label{sec:related_work}
Edge-BehAV draws on five research threads, each corresponding to a component of the proposed offline framework. We first review language-guided navigation, which frames the instruction-following problem, followed by vision-language models for perception and grounding and their edge deployment, which together enable onboard language understanding and semantic goal grounding. We then examine LiDAR-camera fusion and model predictive control, which underpin the geometric goal localization and behavior-guided planning that distinguish our fully offline system from prior cloud-dependent approaches.

\subsection{Language-Guided Robot Navigation}

The grounding of natural language in robot behavior has progressed from template-based to fully neural systems. Early Vision-and-Language Navigation benchmarks~\cite{8578485,gu-etal-2022-vision} showed feasibility in constrained indoor settings with discrete actions but did not generalize outdoors. SayCan~\cite{ahn2022saycan} scores language-model affordances against executable skills, and LM-Nav~\cite{shah2023lmnav} composes pre-trained language, vision, and action models to follow instructions in real environments — both, however, require continuous cloud API access. Foundation-model navigation policies such as ViNT~\cite{shah2023vint} and NoMaD~\cite{10610665} learn broad traversability priors but do not encode arbitrary natural-language behavioral rules, while CoNVOI~\cite{10802716} couples VLMs with a planner for indoor and outdoor scenes, but relies on remote inference. BehAV~\cite{weerakoon2025behav} is the most closely related prior work; Edge-BehAV extends its architecture by replacing every cloud-dependent component with an offline alternative and adding geometric goal localization.

\subsection{Vision-Language Models for Perception and Grounding}

CLIP~\cite{radford2021clip} established open-vocabulary understanding through joint image-text embedding, and CLIPSeg~\cite{luddecke2022clipseg} extended it to dense per-pixel segmentation — the approach BehAV uses for its cost map, which we retain. For open-vocabulary detection, Grounding DINO~\cite{liu2023groundingdino} and Florence-2~\cite{xiao2024florence2} are the state of the art; we select Florence-2 for its unified multi-task architecture, favorable embedded-GPU latency, and strong zero-shot grounding on nuImages~\cite{nuscenes} (Section~\ref{sec:results}). Mobile-SAM~\cite{zhang2023mobilesam} gives prompt-based segmentation at far lower cost than the original SAM~\cite{kirillov2023sam}, practical on the Orin NX. Together, they realize Search-then-Segment, confining detection to the queried target region rather than the whole image.

\subsection{Edge Deployment of LLMs and VLMs}

Post-training quantization~\cite{frantar2023gptq} lets large models run on resource-constrained hardware by reducing weights to INT4/INT8 with minimal accuracy loss. Small yet capable families — Qwen2.5~\cite{qwen2025report}, Gemma3~\cite{google2025gemma3}, Phi-3~\cite{abdin2024phi3} — show that 1.5–7B models can approach cloud-API quality on structured reasoning and instruction following within the 4--8\,GB VRAM envelope of embedded accelerators such as the Jetson Orin NX, deployed here via the Ollama framework. We benchmark this capability specifically for navigation instruction parsing across model families and realistic on-robot platforms.

\begin{figure*}[t]
    \centering
    \includegraphics[width=0.98\textwidth]{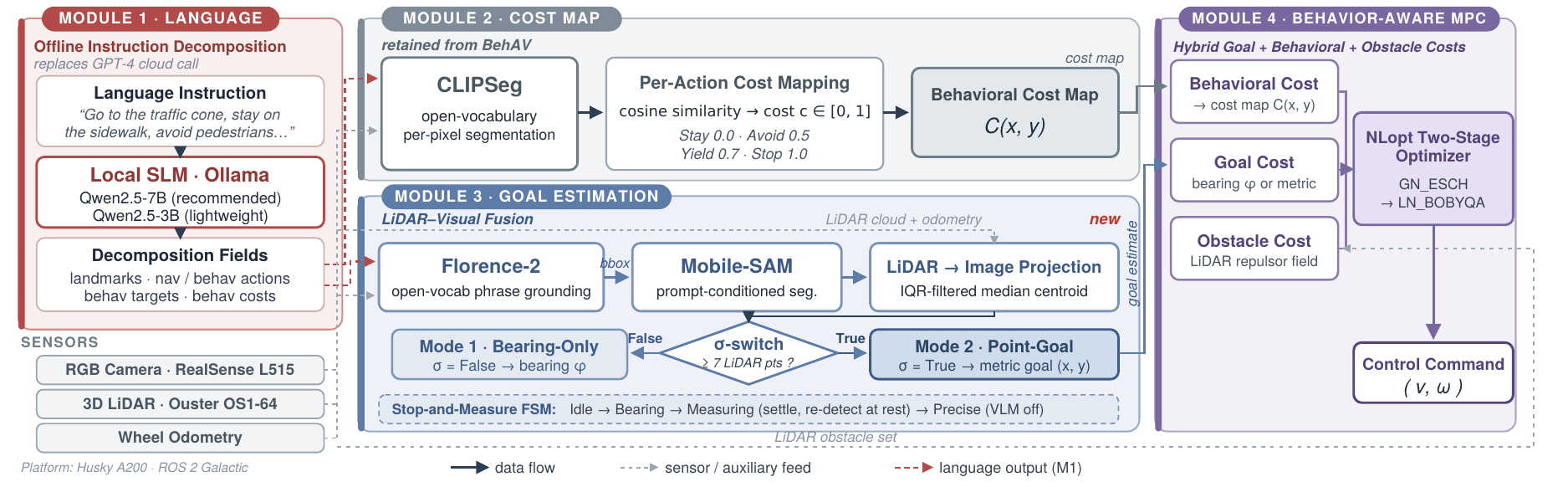}
    \caption{Edge-BehAV system architecture. All four modules run entirely on the robot's onboard computer without internet connectivity. Module~1 replaces the cloud GPT-4 call with a quantized on-device SLM. Module~2 retains the CLIPSeg behavioural cost map from BehAV~\cite{weerakoon2025behav}. Module~3 introduces LiDAR-visual fusion with a Florence-2\,+\,Mobile-SAM Search-then-Segment pipeline and a $\sigma$-switch for metric versus bearing-only goal estimation. Module~4 solves the hybrid MPC objective with a two-stage NLopt optimiser.}
    \label{fig:sysarc}
    \vspace{-1.5em}
\end{figure*}

\subsection{LiDAR-Camera Sensor Fusion for Robot Navigation}

Projecting 3D LiDAR returns onto 2D image features is well established for associating metric depth with semantics. PointPainting~\cite{vora2020pointpainting} paints point clouds with image-space semantic scores to improve 3D detection; Frustum PointNets~\cite{qi2018frustum} uses 2D boxes to define frustum regions in LiDAR space for precise localization; and OpenScene~\cite{peng2023openscene} distils CLIP features into a neural point cloud for voxel-level language queries. Most language-guided navigation systems, however, treat LiDAR and camera independently, LiDAR for occupancy, camera for semantics, without fusing them for metric goal localization. Edge-BehAV closes this gap by gating the LiDAR cloud with the VLM segmentation mask and computing a robust (median) centroid via IQR-based outlier rejection, giving a 0.20\,m mean absolute error in goal localization (Section~\ref{sec:results}).

\subsection{Model Predictive Control for Behavioural Navigation}

MPC planners suit behavior-aware navigation, optimizing over a receding horizon while accommodating heterogeneous cost terms~\cite{williams2018mppi}. BehAV's unconstrained MPC~\cite{weerakoon2025behav} parameterizes candidate trajectories egocentrically (a polar distance, two heading angles, a maximum speed) and solves them with a gradient-free two-stage NLopt optimizer (global GN\_ESCH then local LN\_BOBYQA). Edge-BehAV keeps this structure and adds a hybrid goal cost spanning bearing-only and point-goal modes, a distance-scheduled obstacle weight that prevents goal-approach stalling, and a reduced behavioral-cost decay for earlier avoidance.


\section{Proposed Method}\label{sec:method}

Edge-BehAV retains BehAV's high-level four-module architecture while redesigning the cloud-dependent language understanding and semantic perception modules, introducing sensor-grounded metric goal localization, and preserving the original behavior-guided planning framework. Fig.~\ref{fig:sysarc} highlights the inherited BehAV architecture (gray) together with the redesigned modules proposed in this work.

\subsection{Offline Language Decomposition}

BehAV performs instruction decomposition using GPT-4 through cloud APIs. To eliminate cloud dependency while preserving BehAV's original instruction interface, Edge-BehAV replaces this module (\textit{Module 1} of Fig.~\ref{fig:sysarc}) with an onboard SLM that generates the same structured navigation representations for downstream modules. 
The selected SLM is deployed locally using the Ollama inference engine with the default distribution weights for each tag (e.g. qwen2.5:3b), enabling efficient onboard inference on embedded platforms.
For each navigation instruction, the model is prompted to generate five structured fields: \textit{landmarks(destination)}, \textit{navigation actions}, \textit{behavioral actions}, \textit{behavioral targets}, and \textit{behavioral costs}, where the last specifies a per-target cost used to construct the CLIPSeg behavioral cost map in later part. The SLM runs once per mission at startup and no connectivity is required.

\subsection{Target-Centric Perception Pipeline}
In the\textit{Module 3} of Fig.~\ref{fig:sysarc}, 
BehAV segments the entire image with FastSAM~\cite{zhao2023fast} and then queries a cloud VLM to identify the goal mask, following a Segment-All-then-Classify pipeline, which is computationally expensive and time-consuming. In contrast, Edge-BehAV uses Search-then-Segment strategy that first localizes the target and then segments only the corresponding image region. Specifically, Florence-2~\cite{xiao2024florence2} takes the SLM-extracted goal description and returns a focused bounding box via zero-shot open-vocabulary detection, and Mobile-SAM~\cite{zhang2023mobilesam} generates a high-fidelity mask within that box. This confines segmentation to the target, avoids cloud queries, and processes only the target region rather than the whole scene.

\subsection{LiDAR-Visual Goal Estimation}
Regarding the goal estimation in \textit{Module 3} of Fig.~\ref{fig:sysarc}, unlike the original BehAV, Edge-BehAV fuse LiDAR geometric information, rather than relying solely on visual reasoning.
Let the Mobile-SAM mask $\mathcal{M} \subset \mathbb{R}^{H \times W}$ define the set of pixels attributed to the goal object. LiDAR points $\mathbf{p}_i = (x_i, y_i, z_i)$ in the sensor frame are projected onto the image plane via the extrinsic transformation $\mathbf{T}_{\text{lidar}\to\text{cam}}$ and the camera intrinsic matrix $K$:
\begin{equation}
    \begin{bmatrix} u_i \\ v_i \\ 1 \end{bmatrix}
    \sim K\,\mathbf{T}_{\text{lidar}\to\text{cam}}\,\begin{bmatrix} x_i \\ y_i \\ z_i \\ 1 \end{bmatrix}.
    \label{eq:projection}
\end{equation}
Only points with $z_i > 0$ and $(u_i, v_i) \in \mathcal{M}$ are retained. To reject ground-plane contamination and noise, an IQR filter is applied to the planar distances $d_i = \|(x_i, y_i)\|_2$:
\begin{equation}
    \mathcal{P}_{\text{valid}} = \bigl\{i : Q_1 - 0.5\,\mathrm{IQR} \leq d_i \leq Q_3 + 0.5\,\mathrm{IQR}\bigr\},
    \label{eq:iqr}
\end{equation}
where $Q_1$, $Q_3$ are the 25th and 75th percentiles and $\mathrm{IQR} = Q_3 - Q_1$. If $|\mathcal{P}_{\text{valid}}| \geq N_{\text{pts}} = 7$, the goal range and bearing in the robot frame follow from the robust (median) centroid $\tilde{x} = \mathrm{median}(x_{\text{valid}})$, $\tilde{y} = \mathrm{median}(y_{\text{valid}})$ of the retained points, using the median in place of the mean for outlier robustness:
\begin{equation}
    r = \sqrt{\tilde{x}^2 + \tilde{y}^2}, \qquad
    \psi = \mathrm{atan2}(\tilde{y},\, \tilde{x}).
    \label{eq:centroid}
\end{equation}
The world-frame goal is then latched as:
\begin{equation}
    (g_x,\, g_y) = \bigl(x_r + r\cos(\theta_r + \psi),\; y_r + r\sin(\theta_r + \psi)\bigr),
    \label{eq:goal_latch}
\end{equation}
with $(x_r, y_r, \theta_r)$ the robot pose from odometry.

\subsection{Hybrid Goal Cost and Mode Switching}
In \textit{Module 3} of the Fig.~\ref{fig:sysarc},
when the goal is too far or returns insufficient LiDAR points ($|\mathcal{P}_{\text{valid}}| < N_{\text{pts}} = 7$), no metric goal can be latched via Eq.~\eqref{eq:goal_latch}. Rather than halting, Edge-BehAV enters a \textit{bearing-only mode} that keeps advancing toward the target direction derived from the 2D visual detection.

\subsubsection{Bearing Estimation from Pixel Coordinates}

Given the horizontal centre $u_c = (u_{\min} + u_{\max})/2$ of the Florence-2 box, the target ray in the camera optical frame is:
\begin{equation}
    \mathbf{d}_{\text{cam}} = \begin{bmatrix}(u_c - c_x)/f_x \\ 0 \\ 1\end{bmatrix},
    \label{eq:ray_cam}
\end{equation}
where $f_x$ and $c_x$ are the horizontal focal length and principal-point offset from $K$. Rotating into the robot body frame, $\mathbf{d}_{\text{robot}} = \mathbf{R}_{\text{cam}\to\text{robot}}\,\mathbf{d}_{\text{cam}}$, yields the robot-relative bearing $\beta = \mathrm{atan2}(d_{\text{robot},y}, d_{\text{robot},x})$ and the target heading $\phi_{\text{target}} = \theta_r + \beta$ in the odometry frame.

\subsubsection{Hybrid MPC Goal Cost}

Let $\sigma \in \{\text{True},\text{False}\}$ denote the switching flag: $\sigma = \text{True}$ when a valid metric goal is latched (Point-Goal Mode, $|\mathcal{P}_{\text{valid}}| \geq N_{\text{pts}}$), and $\sigma = \text{False}$ otherwise (Bearing Mode). The goal cost term in the MPC objective function is:
\begin{equation}
    \psi_{\text{goal}} =
    \begin{cases}
        \displaystyle\sum_{t=1}^{T} \frac{d_t}{d_{\text{tot}}}
            & \text{if } \sigma = \text{True},\\[8pt]
        \displaystyle\sum_{t=1}^{T}\!\left(
            w_\alpha\,\bigl|\theta_t - \phi_{\text{target}}\bigr|
            + w_v\,(V_{\text{max}} - v_t)
        \right)
            & \text{if } \sigma = \text{False},
    \end{cases}
    \label{eq:goal_cost}
\end{equation}
where $d_t = \|(x_t - g_x, y_t - g_y)\|_2$ is the distance from waypoint $t$ to the latched goal, $d_{\text{tot}}$ the initial robot-to-goal distance, $\theta_t$ and $v_t$ the trajectory heading and speed at step $t$, $V_{\text{max}}$ the maximum commanded speed, and $w_\alpha = 1.5$, $w_v = 0.5$ tuned weights. In Bearing Mode the first term penalises heading misalignment with $\phi_{\text{target}}$ and the second penalises insufficient forward progress, preventing the robot from turning in place.

\subsubsection{Stop-and-Measure State Machine}

Latching the goal while the robot moves introduces motion-induced blur and temporal desynchronisation between the RGB frame, LiDAR scan, and odometry. To eliminate both, Edge-BehAV uses a four-state machine \textsc{Idle}~(0) $\to$ \textsc{Bearing}~(1) $\to$ \textsc{Measuring}~(3) $\to$ \textsc{Precise}~(2). When a Florence-2 detection is obtained in Bearing Mode with $|\mathcal{P}_{\text{valid}}| \geq N_{\text{pts}}$, the system enters \textsc{Measuring}: the VLM thread is disabled and the planner publishes zero velocity to stop the robot. After a settle period the full pipeline (Florence-2~$\to$~Mobile-SAM~$\to$ LiDAR projection) is re-run on a \emph{fresh} image and scan at the stationary pose; because the image, cloud, and odometry are consistent at rest, the goal is latched via Eq.~\eqref{eq:goal_latch} and the system transitions to \textsc{Precise}, permanently disabling the VLM. If the stationary re-detection fails (target out of view or too few returns), it reverts to \textsc{Bearing} for another approach cycle.

\subsubsection{Full MPC Objective and Weight Schedule}

The complete MPC objective is:
\begin{equation}
    J = w_{\text{goal}}\,\psi_{\text{goal}} + w_{\text{behav}}\,\psi_{\text{behav}} + w_{\text{obs}}\,\psi_{\text{obs}},
    \label{eq:total_cost}
\end{equation}
with $w_{\text{goal}} = 2.0$ and $w_{\text{behav}} = 8.0$.

The behavioral cost projects each waypoint into the CLIPSeg cost map with an exponential distance decay:
\begin{equation}
    \psi_{\text{behav}} = \max_{t=1}^{T}\; C_{\text{behav}}(x_t,y_t)\;\exp(-\lambda_b\,\|{(x_t,y_t) - \mathbf{p}_r}\|_2),
    \label{eq:behav_cost}
\end{equation}
where $\mathbf{p}_r$ is the robot position and $\lambda_b = 0.3$. 
The obstacle weight is scheduled in Point-Goal Mode to stop goal-attraction and obstacle-repulsion competing on final approach:
\begin{equation}
    w_{\text{obs}} = 3.0 \times \min\!\left(1.0,\;\frac{d_{\text{goal}}}{2.0}\right),
    \label{eq:w_obs}
\end{equation}
ramping $w_{\text{obs}}$ from zero at arrival to 3.0 beyond 2\,m. In Mode~2, LiDAR returns within 1.2\,m of the latched goal are also excluded from the obstacle set, preventing the target's own surface from generating repulsive cost.


\begin{figure}[h]
    \centering
    \includegraphics[width=0.4\textwidth]{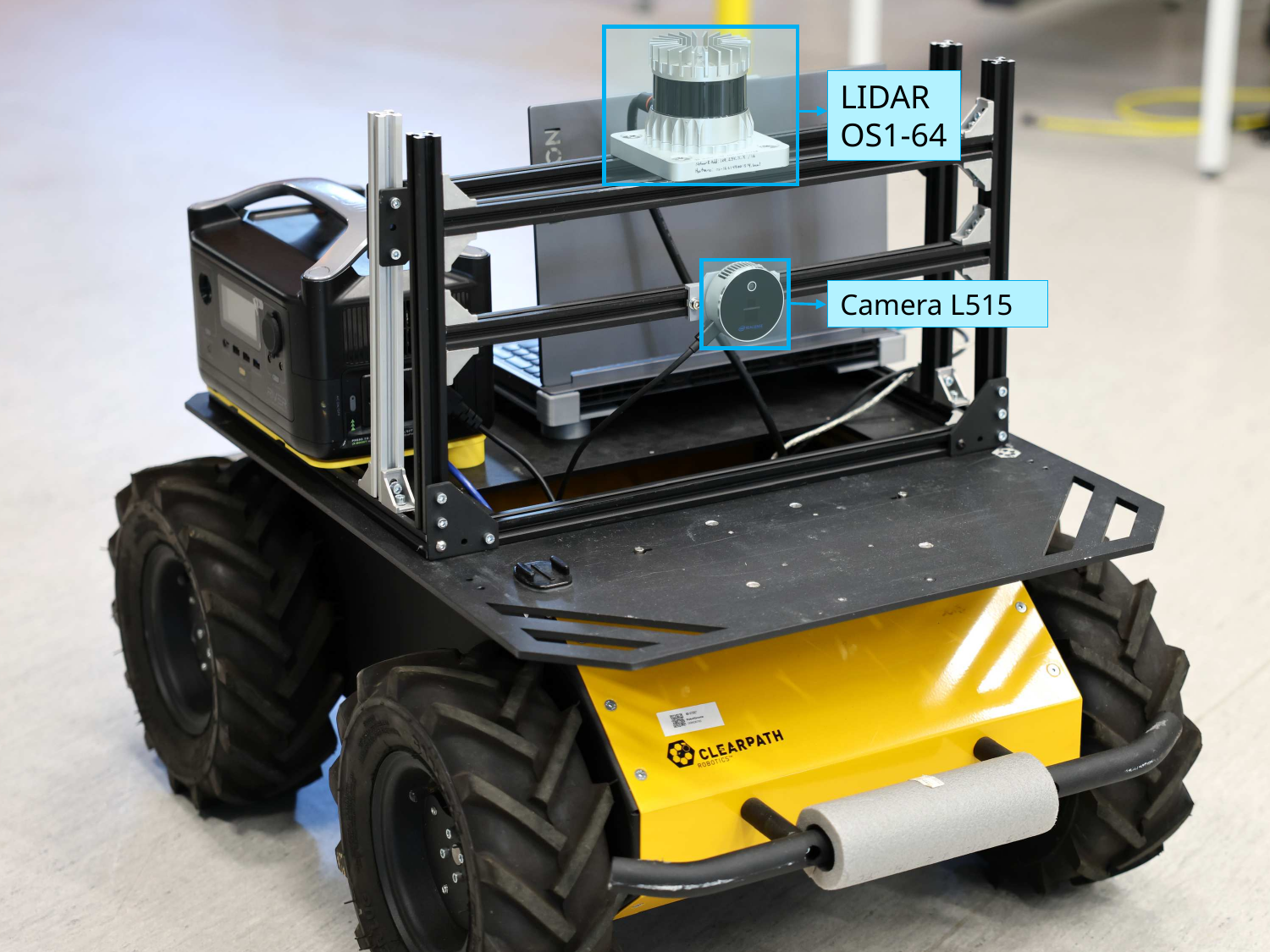}
    \caption{Customized Clearpath Husky\,A200 with an Ouster OS1-64 3D LiDAR and an Intel RealSense L515 depth camera.}
    \label{fig:roboticplatform}
    \vspace{-1.5em}
\end{figure}

\section{Experimental Setting}\label{sec:experiment}

\subsection{Hardware Platforms}
To evaluate deployment across representative robotic computing platforms,
experiments span three configurations, from consumer GPU to embedded targets. Quality benchmarks run on a laptop with an NVIDIA RTX 4070 (8\,GB VRAM); embedded latency is verified on two NVIDIA Jetson platforms, the Orin NX 16\,GB and the Thor. Full-system integration runs on a Clearpath Husky\,A200 with an Ouster OS1-64 3D LiDAR and an Intel RealSense L515 depth camera under ROS\,2 Galactic. The robotic platform is shown in Fig.~\ref{fig:roboticplatform}.

\subsection{Language Decomposition Benchmark}

To evaluate offline language decomposition, we construct a benchmark of 350 natural-language navigation instructions spanning seven equal-size semantic categories (50 prompts each): \textit{Surface/Terrain}, \textit{Human Interaction}, \textit{Static Obstacles}, \textit{Compound Instructions}, \textit{Contextual/Adversarial}, \textit{Off-Road \& Natural Terrain}, and \textit{Wet \& Winter Terrain}. To enable reproducible comparison as new models appear, the annotated benchmark is publicly available at \url{https://github.com/aliiisa1375/Edge_Behav}.

We evaluate 17 local small language models (SLMs) from seven families: Qwen2.5, Qwen2.5-VL, Qwen3, Qwen3.5, Gemma3, Gemma3n, and Gemma4; together with four cloud APIs: GPT-5.5, GPT-5.4-mini, GPT-5.4-nano, and GPT-4, the last standing in for the cloud decomposition backend of the original BehAV system~\cite{weerakoon2025behav}. All local models served via Ollama; all 21 returned usable output on all 350 instructions. Cloud models use BehAV's original prompt but a type-tolerant field extractor, as its list-only one discarded 63\% of GPT-4's values (0.28$\rightarrow$0.512).

Each instruction is submitted with an identical structured prompt requesting four fields of: \textit{landmarks}, \textit{navigation actions}, \textit{behavioral actions}, \textit{behavioral targets}, plus a fifth \textit{behavioral costs} field (a real-valued compliance penalty per target) extracted separately. Quality is measured against a manually curated ground-truth (GT): each instruction was independently annotated by the authors into a reference JSON decomposition, with behavioral costs following a four-level mapping — \textit{stay-on} 0.0, \textit{avoidance} 0.5, \textit{yielding} 0.7, \textit{stopping} 1.0.
This fixed, inspectable benchmark ensures scores reflect agreement with human-verified annotations rather than any single model's live output.

\subsection{Landmark Detection Benchmark}

The goal detection module is evaluated on a curated subset of $n=922$ nuImages~\cite{nuscenes} images covering seven outdoor landmark classes, with ground-truth boxes for detection recall and localization accuracy. Three methods are compared: FastSAM + GPT-4o (BehAV baseline~\cite{weerakoon2025behav}), FastSAM + GPT-5.5 (current cloud SOTA), and Florence-2 open-vocabulary detection (ours).

\subsection{Distance Estimation Benchmark}

The distance is evaluated on the Husky\,A200 against four landmark categories (traffic cone, car, bicycle, dumpster), each placed at 2, 5, 10, and 20\,m with tape-measured ground truth. The LiDAR--VLM fusion pipeline (Florence-2 box $\to$ Mobile-SAM mask $\to$ OS1-64 centroid) is compared against the visual-heuristic baseline of BehAV~\cite{weerakoon2025behav}, which infers depth from the 2D projection without metric sensing. Absolute error $|e| = |\hat{d} - d_{\text{true}}|$ is reported per trial and Mean Absolute Error(MAE) over all valid measurements across the trials.

\subsection{Full-System Closed-Loop Benchmark}

To verify that the three offline modules operate jointly on a physical robot, the full pipeline is evaluated closed-loop on the Husky\,A200 across four outdoor scenarios (\textit{Scn.1 -- 4}), each repeated for $N=8$ trials from a fixed start pose. \textit{Scn.1} (dynamic human avoidance) drives to goal 
while avoiding people on the path. 
\textit{Scn.2} (bearing-to-metric transition) approaches a traffic cone while avoiding a water puddle, exercising the handover from bearing-only tracking to a latched metric goal. \textit{Scn.3} (goal disambiguation) drives to a green dumpster placed among a visually similar look-alike at 14\,m while avoiding a traffic cone;
\textit{Scn.4} pushes the same disambiguation task to long range (26\,m) from a different viewpoint with two traffic cones on the approach. 

For \textit{Scn.3} and \textit{Scn.4} the original BehAV system~\cite{weerakoon2025behav} is additionally run closed-loop as a baseline, using the authors' released planner. Because that planner takes its goal by manual operator entry from a one-shot GPT distance estimate and lacks a LiDAR obstacle term, the comparison is illustrative rather than perfectly controlled; the strictly controlled, same-input comparisons remain those of Tables~\ref{tab:slm_main}, \ref{tab:detection}, and~\ref{tab:distance}.

\subsection{Evaluation Metrics}

\textit{Language decomposition quality} is evaluated using set-level Macro-F1: each of the four fields (\textit{landmarks}, \textit{navigation actions}, \textit{behavioral actions}, \textit{behavioral targets}) is treated as a set-matching problem (output tokens vs.\ GT tokens by token-level intersection) and the field F1s are averaged, with a 95\% bootstrap confidence interval (2000 resamples over the 350 per-instruction scores). Median latency per call and throughput (tokens/s) are secondary metrics.

\textit{Landmark detection quality} is F-score and mean pixel error (Euclidean distance from predicted to ground-truth bounding-box centroid) over all detected images. \textit{Distance estimation quality} is absolute error $|e|$ per measurement and MAE over valid trials; a trial is invalid (`---') when the LiDAR mask returns fewer points than the minimum for a reliable centroid.

\textit{Closed-loop quality} uses five per-trial metrics averaged over each scenario's eight trials. \textit{Success Rate} (SR) is the fraction of trials reaching the goal within tolerance ($1.0\pm0.3$\,m) with no collision and the behavioral rule satisfied. \textit{Goal Heading Error} (GHE) is the whole-path mean angle between robot heading and line-of-sight bearing to the latched goal. \textit{Goal Distance Error} (GDE)\,$=|\hat{d}-d_{\text{true}}|$ is the metric-estimate error at first goal lock relative to tape-measured ground truth. \textit{Fr\'echet distance} is the discrete Fr\'echet distance to a human-teleoperated reference run per scenario, and \textit{Behavior Following Accuracy} (BFA) is the arc-length fraction complying with the behavioral rule, scored by point-in-polygon test against a pre-surveyed map of forbidden regions (water-puddle rim in \textit{Scn.2}; keep-out boxes of half-width $1.0$\,m around each surveyed cone in  \textit{Scn.3} and \textit{Scn.4}), both as defined by BehAV~\cite{weerakoon2025behav}. BFA is reported only where the forbidden geometry is static and surveyable, so it is omitted for \textit{Scn.1}'s unsurveyed human obstacles.


\section{Experimental Results}\label{sec:results}

\subsection{Offline Language Decomposition}

Table~\ref{tab:slm_main} reports Macro-F1, 95\% bootstrap confidence intervals, per-field F1, and median latency for a representative set of models (all 21 in the supplementary data), scored against the human-annotated ground truth (GT), and Figure~\ref{fig:pareto} plots the quality--latency trade-off. Three findings stand out. \emph{(i)}~The strongest offline model reaches cloud-level quality: Qwen2.5-7B attains the top offline Macro-F1 of 0.775, statistically on par with the best cloud API GPT-5.5 (0.765; the gap lies within overlapping confidence intervals) while running roughly $9\times$ faster and fully offline. \emph{(ii)}~The GPT-4 backend of the original BehAV system~\cite{weerakoon2025behav} reaches 0.512, below all but one offline model, quantifying the accuracy gap that motivates a purpose-built offline decomposer. \emph{(iii)}~The text-only LLMs outperform the vision-language variants on this decomposition task, so a VLM is warranted only when the same model must additionally ground the instruction against imagery. We therefore adopt Qwen2.5-7B as the recommended deployment model, with Qwen2.5-3B as a lightweight alternative (0.718 Macro-F1 at 0.53\,s within a 4\,GB VRAM budget).

\begin{table*}[t]
\centering

\begin{minipage}[t]{0.55\textwidth}
\centering
\caption{SLM decomposition on 350 instructions (7 categories) vs.\ the human-annotated GT. Macro-F1 = mean set-level F1 over four fields; CI$_{95}$ = 95\% bootstrap CI; Lmk-/Nav-F1 = landmark/navigation-action fields; latency = median per call. Bold = offline Pareto-optimal; $\star$ = recommended model.}
\label{tab:slm_main}

\scriptsize
\setlength{\tabcolsep}{6.8pt}
\renewcommand{\arraystretch}{1.2}

\begin{tabular}{@{}llccccc@{}}
\toprule
Model & Type & Macro-F1 & CI$_{95}$ & Lmk-F1 & Nav-F1 & Lat.\ (s) \\
\midrule
\multicolumn{7}{@{}l}{\textit{Cloud APIs (comparison only --- require internet)}} \\
GPT-5.5
& cloud & 0.765 & [0.753,\,0.776] & 0.644 & 0.630 & 9.95 \\
GPT-5.4-mini
& cloud & 0.700 & [0.684,\,0.715] & 0.537 & 0.614 & 2.19 \\
GPT-4~\cite{weerakoon2025behav}
& cloud & 0.512 & [0.486,\,0.540] & 0.417 & 0.389 & 4.05 \\
\midrule

\multicolumn{7}{@{}l}{\textit{Local language models (LLMs) --- offline}} \\
\textbf{Qwen2.5-7B}$^{\star}$
& LLM & \textbf{0.775} & [0.756,\,0.793] & 0.750 & 0.627 & 1.13 \\
\textbf{Qwen2.5-3B}
& LLM & \textbf{0.718} & [0.697,\,0.737] & 0.756 & 0.618 & 0.53 \\
Gemma4-E4B
& LLM & 0.716 & [0.700,\,0.732] & 0.772 & 0.590 & 2.77 \\
Gemma3-12B
& LLM & 0.683 & [0.664,\,0.703] & 0.629 & 0.663 & 2.95 \\
\textbf{Qwen3-1.7B}
& LLM & \textbf{0.596} & [0.582,\,0.610] & 0.621 & 0.409 & 0.46 \\
\textbf{Qwen2.5-1.5B}
& LLM & \textbf{0.325} & [0.303,\,0.348] & 0.118 & 0.372 & 0.33 \\
\midrule

\multicolumn{7}{@{}l}{\textit{Local vision-language models (VLMs) --- offline}} \\
Qwen2.5-VL-7B
& VLM & 0.662 & [0.643,\,0.679] & 0.684 & 0.455 & 4.58 \\
Qwen2.5-VL-3B
& VLM & 0.567 & [0.548,\,0.586] & 0.289 & 0.412 & 2.21 \\
\bottomrule
\end{tabular}
\end{minipage}
\hspace{1em}
\begin{minipage}[t]{0.4\textwidth}
\centering

\caption{Cross-platform Macro-F1 and latency for identical INT4 checkpoints (350 instructions, vs.\ GT). $\star$ = recommended model.}
\label{tab:cross_platform}

\scriptsize
\setlength{\tabcolsep}{11.0pt}
\renewcommand{\arraystretch}{1.02}

\begin{tabular}{@{}lcc@{}}
\toprule
Model & Macro-F1 & Latency (s) \\
\midrule
\textbf{Qwen2.5-7B}$^{\star}$
& \textbf{0.775 / 0.743 / 0.744}
& 1.13 / 5.27 / 1.72 \\
Qwen2.5-3B
& 0.718 / 0.726 / 0.727
& 0.53 / 2.66 / 0.99 \\
Qwen2.5-VL-7B
& 0.662 / 0.656 / 0.661
& 4.58 / 5.40 / 1.59 \\
Gemma3n-E4B
& 0.647 / 0.662 / 0.656
& 1.70 / 5.05 / 2.15 \\
Qwen3-4B
& 0.568 / 0.567 / 0.564
& 0.80 / 3.70 / 1.37 \\
Qwen2.5-VL-3B
& 0.567 / 0.575 / 0.579
& 2.21 / 2.88 / 1.02 \\
Gemma3-4B
& 0.554 / 0.582 / 0.595
& 0.84 / 4.20 / 1.55 \\
Qwen2.5-1.5B
& 0.325 / 0.326 / 0.327
& 0.33 / 1.71 / 0.64 \\
\bottomrule
\end{tabular}

\vspace{1mm}

{\scriptsize
Macro-F1 and latency are reported as
\textbf{Laptop / Orin NX / Thor}.}

\vspace{3mm}

\caption{Landmark detection on the nuImages subset ($n=922$, 7 classes). Time = end-to-end per image.}
\label{tab:detection}

\scriptsize
\setlength{\tabcolsep}{1.8pt}
\renewcommand{\arraystretch}{1.05}

\resizebox{\linewidth}{!}{%
\begin{tabular}{@{}lccc@{}}
\toprule
Method
& Pixel Error$\downarrow$
& F-Score$\uparrow$
& Time (ms)$\downarrow$ \\
\midrule
FastSAM + GPT-4o~\cite{weerakoon2025behav}
& 99.10\,$\pm$\,148.87
& 0.904
& 3197.8\,$\pm$\,1363.8 \\
FastSAM + GPT-5.5
& 44.97\,$\pm$\,101.40
& 0.768
& 4861.1\,$\pm$\,2661.0 \\
Florence-2 (ours)
& \textbf{38.02\,$\pm$\,130.03}
& \textbf{0.974}
& \textbf{306.9\,$\pm$\,14.3} \\
\bottomrule
\end{tabular}%
}

\end{minipage}
\vspace{-1.0em}
\end{table*}

As Fig.~\ref{fig:pareto} shows, four offline models form the quality--latency Pareto frontier: Qwen2.5-7B$^\star$ anchors the high-quality end, matching every cloud API in quality while dominating on latency, and Qwen2.5-3B sits at the knee as the lightweight alternative, capturing 93\% of the top offline quality at less than half the latency.



\definecolor{ParetoBlue}{HTML}{2E7EBB}
\definecolor{DomGray}{HTML}{8C8C8C}
\definecolor{CloudRed}{HTML}{C0392B}

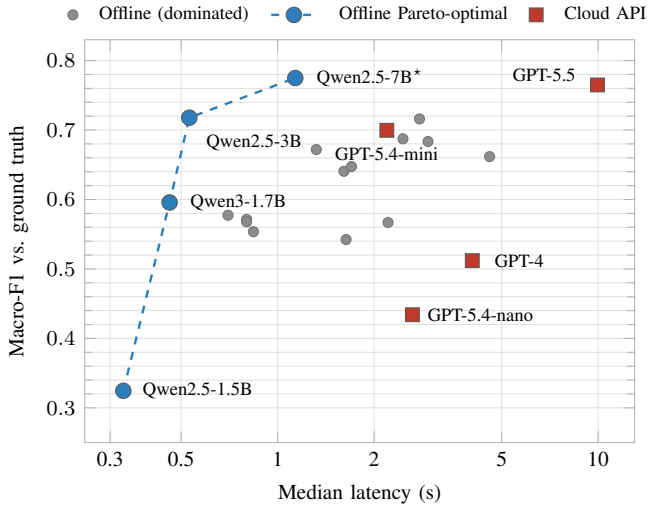
\begin{figure}[!htb]
\centering
\begin{tikzpicture}
\begin{axis}[
    width=\linewidth,
    height=0.78\linewidth,
    xmode=log,
    log basis x=10,
    xlabel={Median latency (s)},
    ylabel={Macro-F1 vs.\ ground truth},
    xlabel style={font=\footnotesize},
    ylabel style={font=\footnotesize},
    tick label style={font=\footnotesize},
    xmin=0.25, xmax=13,
    ymin=0.25, ymax=0.83,
    xtick={0.3,0.5,1,2,5,10},
    xticklabels={0.3,0.5,1,2,5,10},
    ytick={0.3,0.4,0.5,0.6,0.7,0.8},
    grid=both,
    grid style={gray!25, line width=0.3pt},
    minor tick num=4,
    axis line style={gray!60},
    tick style={gray!60},
    legend style={
        font=\scriptsize,
        at={(0.5,1.02)}, anchor=south,
        draw=none, fill=none,
        legend columns=3,
        column sep=6pt, inner sep=1pt,
    },
    legend cell align=left,
]

\addplot[only marks, mark=*, mark size=1.9pt,
         mark options={fill=DomGray, draw=black!55, line width=0.3pt}]
    coordinates {
        (0.700,0.5773)  
        (0.800,0.5713)  
        (0.800,0.5678)  
        (0.840,0.5536)  
        (1.320,0.6719)  
        (1.610,0.6406)  
        (1.635,0.5423)  
        (1.700,0.6474)  
        (2.210,0.5669)  
        (2.460,0.6874)  
        (2.770,0.7160)  
        (2.945,0.6835)  
        (4.585,0.6617)  
    };
\addlegendentry{Offline (dominated)}

\addplot[ParetoBlue, dashed, line width=0.9pt,
         mark=*, mark size=2.9pt,
         mark options={solid, fill=ParetoBlue, draw=black!65, line width=0.4pt}]
    coordinates {
        (0.330,0.3246)  
        (0.460,0.5957)  
        (0.530,0.7178)  
        (1.135,0.7749)  
    };
\addlegendentry{Offline Pareto-optimal}

\addplot[only marks, mark=square*, mark size=2.6pt,
         mark options={fill=CloudRed, draw=black!65, line width=0.4pt}]
    coordinates {
        (9.952,0.7649)  
        (2.191,0.6996)  
        (2.633,0.4340)  
        (4.048,0.5119)  
    };
\addlegendentry{Cloud API}

\node[font=\scriptsize, anchor=west]      at (axis cs:1.24,0.7749) {Qwen2.5-7B$^{\star}$};
\node[font=\scriptsize, anchor=north west] at (axis cs:0.56,0.7075) {Qwen2.5-3B};
\node[font=\scriptsize, anchor=west]      at (axis cs:0.50,0.5957) {Qwen3-1.7B};
\node[font=\scriptsize, anchor=west]      at (axis cs:0.355,0.3246) {Qwen2.5-1.5B};
\node[font=\scriptsize, anchor=east]      at (axis cs:9.10,0.7790) {GPT-5.5};
\node[font=\scriptsize, anchor=north]     at (axis cs:2.191,0.6880) {GPT-5.4-mini};
\node[font=\scriptsize, anchor=west]      at (axis cs:2.75,0.4335) {GPT-5.4-nano};
\node[font=\scriptsize, anchor=west]      at (axis cs:4.45,0.5119) {GPT-4};
\end{axis}
\end{tikzpicture}
\caption{Quality--latency Pareto frontier (21 models). Blue = offline Pareto-optimal; grey = offline but dominated; red = cloud APIs. 
}
\label{fig:pareto}
\end{figure}

\subsection{Per-Category and Cross-Platform Analysis}

Fig.~\ref{fig:heatmap} shows per-category Macro-F1 for all models. Averaged over models, the hardest categories are Wet \& Winter Terrain and Contextual/Adversarial, confirming the intended stress-test effect of the two terrain categories; 
Qwen2.5-7B is the most uniform performer (every category between 0.746 and 0.821), and Qwen2.5-3B stays above 0.70 on five of seven categories.



\pgfplotsset{
  colormap={RdYlGn}{
    rgb255(0cm)=(215,48,39)
    rgb255(1cm)=(244,109,67)
    rgb255(2cm)=(253,174,97)
    rgb255(3cm)=(254,224,139)
    rgb255(4cm)=(217,239,139)
    rgb255(5cm)=(166,217,106)
    rgb255(6cm)=(102,189,99)
    rgb255(7cm)=(26,152,80)
  },
}

\begin{figure}[!htb]
\centering
\begin{tikzpicture}
\begin{axis}[
    width=0.64\linewidth,
    height=0.90\linewidth,
    scale only axis,
    enlargelimits=false,
    axis on top,
    xmin=-0.5, xmax=6.5,
    ymin=-0.5, ymax=20.5,
    y dir=reverse,
    xtick={0,1,2,3,4,5,6},
    xticklabels={Compound, Contextual/Adv., Human Interaction,
                 Off-Road \& Natural, Static Obstacles, Surface/Terrain,
                 Wet \& Winter},
    x tick label style={rotate=40, anchor=north east, font=\scriptsize,
                        yshift=1pt, xshift=1pt},
    xtick pos=bottom,
    ytick={0,1,2,3,4,5,6,7,8,9,10,11,12,13,14,15,16,17,18,19,20},
    yticklabels={Qwen2.5-7B, GPT-5.5, Qwen2.5-3B, Gemma4-E4B, GPT-5.4-mini,
                 Qwen3.5-9B, Gemma3-12B, Qwen3-8B, Qwen2.5-VL-7B, Gemma3n-E4B,
                 Qwen3.5-4B, Qwen3-1.7B, Gemma3n-E2B, Qwen3.5-2B, Qwen3-4B,
                 Qwen2.5-VL-3B, Gemma3-4B, Gemma4-E2B, GPT-4,
                 GPT-5.4-nano, Qwen2.5-1.5B},
    y tick label style={font=\scriptsize},
    ytick pos=left,
    tick style={draw=none},
    colorbar,
    colormap name=RdYlGn,
    point meta min=0.10, point meta max=0.85,
    colorbar style={
        width=5pt,
        xshift=-4pt,
        ytick={0.2,0.4,0.6,0.8},
        tick label style={font=\scriptsize, xshift=-2pt},
        ylabel={Macro-F1},
        ylabel style={font=\scriptsize, yshift=6pt},
        tick style={draw=none},
    },
]
\addplot[
    matrix plot*,
    mesh/cols=7,
    mesh/ordering=rowwise,
    point meta=explicit,
    nodes near coords={\pgfmathprintnumber[fixed, zerofill, precision=2]{\pgfplotspointmeta}},
    nodes near coords style={font=\tiny, text=black, anchor=center, yshift=0pt},
] table[meta=v] {
x y v
0 0 0.746
1 0 0.760
2 0 0.800
3 0 0.821
4 0 0.750
5 0 0.763
6 0 0.785
0 1 0.818
1 1 0.751
2 1 0.791
3 1 0.758
4 1 0.740
5 1 0.771
6 1 0.725
0 2 0.697
1 2 0.654
2 2 0.760
3 2 0.708
4 2 0.751
5 2 0.750
6 2 0.704
0 3 0.766
1 3 0.686
2 3 0.697
3 3 0.718
4 3 0.741
5 3 0.730
6 3 0.674
0 4 0.738
1 4 0.679
2 4 0.721
3 4 0.692
4 4 0.681
5 4 0.718
6 4 0.668
0 5 0.669
1 5 0.655
2 5 0.802
3 5 0.692
4 5 0.708
5 5 0.608
6 5 0.678
0 6 0.685
1 6 0.687
2 6 0.743
3 6 0.688
4 6 0.679
5 6 0.667
6 6 0.634
0 7 0.654
1 7 0.654
2 7 0.704
3 7 0.683
4 7 0.699
5 7 0.648
6 7 0.662
0 8 0.525
1 8 0.609
2 8 0.708
3 8 0.699
4 8 0.713
5 8 0.685
6 8 0.693
0 9 0.591
1 9 0.672
2 9 0.702
3 9 0.635
4 9 0.716
5 9 0.600
6 9 0.615
0 10 0.710
1 10 0.585
2 10 0.691
3 10 0.620
4 10 0.669
5 10 0.598
6 10 0.611
0 11 0.616
1 11 0.610
2 11 0.582
3 11 0.558
4 11 0.642
5 11 0.611
6 11 0.551
0 12 0.646
1 12 0.565
2 12 0.551
3 12 0.568
4 12 0.609
5 12 0.552
6 12 0.549
0 13 0.588
1 13 0.557
2 13 0.588
3 13 0.497
4 13 0.678
5 13 0.598
6 13 0.493
0 14 0.643
1 14 0.529
2 14 0.543
3 14 0.564
4 14 0.607
5 14 0.516
6 14 0.573
0 15 0.548
1 15 0.531
2 15 0.564
3 15 0.566
4 15 0.588
5 15 0.585
6 15 0.585
0 16 0.471
1 16 0.561
2 16 0.663
3 16 0.519
4 16 0.649
5 16 0.493
6 16 0.519
0 17 0.442
1 17 0.572
2 17 0.498
3 17 0.571
4 17 0.561
5 17 0.600
6 17 0.552
0 18 0.577
1 18 0.498
2 18 0.453
3 18 0.477
4 18 0.505
5 18 0.544
6 18 0.529
0 19 0.477
1 19 0.418
2 19 0.391
3 19 0.441
4 19 0.399
5 19 0.464
6 19 0.448
0 20 0.490
1 20 0.292
2 20 0.351
3 20 0.261
4 20 0.320
5 20 0.266
6 20 0.292
};
\end{axis}
\end{tikzpicture}
\caption{Per-category Macro-F1 vs.\ GT. Rows are the 21 models sorted by overall Macro-F1; columns are the seven instruction categories.}
\label{fig:heatmap}
\vspace{-0.5em}
\end{figure}

Table~\ref{tab:cross_platform} reports Macro-F1 and median latency across the laptop, Jetson Orin NX, and Jetson Thor. Because quantization fixes the weights, quality is essentially platform-independent: over the eight common models the per-model Macro-F1 spread is at most 0.041 (mean 0.015), and Qwen2.5-7B varies by only 0.032 (0.775/0.743/0.744). Latency is platform-dependent, rising roughly $5\times$ from laptop to Orin NX for sub-3B models, with the Thor recovering most of the gap, but even the 7B model completes in 5.27\,s on the Orin NX, acceptable for one-time instruction parsing. Model selection performed offline on a laptop therefore transfers directly to embedded hardware.


\subsection{Offline Landmark Detection}

Table~\ref{tab:detection} compares Florence-2 against the BehAV cloud baseline (FastSAM\,+\,GPT-4o) and current cloud SOTA (FastSAM\,+\,GPT-5.5) on the 922-image nuImages subset. Florence-2 reaches an F-score of 0.974 at 38.02\,px mean pixel error, improving on both baselines on every metric while running offline at 307\,ms — a $10.4\times$ speedup over GPT-4o and $15.8\times$ over GPT-5.5, which despite better pixel localisation than GPT-4o (44.97 vs.\ 99.10\,px) yields a lower F-score (0.768). This is a system-level comparison, a discriminative detector against a segment-then-classify pipeline, not a controlled detector swap; the offline Search-then-Segment pipeline is both more accurate and more efficient than BehAV's cloud Segment-All pipeline~\cite{weerakoon2025behav}. The pixel-error distribution is heavy-tailed (std 130.03\,px $\gg$ mean 38.02\,px), driven by a few large-error frames rather than systematic bias.



\begin{table}[h]
\centering
\caption{Absolute distance-estimation error on the Husky\,A200.
$|e|$ denotes the absolute error relative to tape-measured ground truth;
`---' indicates too few LiDAR returns for reliable centroid estimation.}
\label{tab:distance}

\scriptsize
\setlength{\tabcolsep}{12pt}
\renewcommand{\arraystretch}{1.28}

\begin{tabular}{ccc}
\toprule
\multirow{2}{*}{$d_{\text{true}}$ (m)}
& \multicolumn{2}{c}{Object order: (Cone, Car, Bike, Dumpster)} \\
\cmidrule(lr){2-3}
& Edge-BehAV $|e|$ (m)
& BehAV $|e|$ (m) \\
\midrule

2
& (0.13,\;0.17,\;0.36,\;0.01)
& (0.00,\;1.00,\;1.00,\;0.95) \\

5
& (0.11,\;0.31,\;0.45,\;0.00)
& (1.00,\;2.00,\;2.00,\;0.93) \\

10
& (---,\;0.13,\;0.39,\;0.01)
& (0.00,\;1.00,\;5.00,\;1.90) \\

20
& (---,\;0.45,\;0.20,\;0.10)
& (2.00,\;1.00,\;8.00,\;5.00) \\

\midrule
\textbf{MAE}
& \textbf{0.20}
& \textbf{2.05} \\
\bottomrule
\end{tabular}
\vspace{-2.5em}
\end{table}

\subsection{Distance Estimation}

Table~\ref{tab:distance} reports metric distance estimation for four object categories at four distances. The LiDAR--VLM fusion pipeline attains an MAE of 0.20\,m over all valid measurements versus 2.05\,m for the BehAV visual heuristic~\cite{weerakoon2025behav}, a $10.3\times$ improvement, with worst-case error below 0.45\,m over the full 2–20\,m range. For the traffic cone, insufficient LiDAR returns at 10 and 20\,m prevent centroid extraction: a density limit of the OS1-64 on small low-reflectance targets at range, not a detection failure. These `---' cells are exactly the regime the bearing-only fallback covers, the system tracks the visual bearing until close enough for a metric lock 
as exercised end-to-end in \textit{Scn.4} (Table~\ref{tab:closed_loop}). The heuristic's largest errors (8.00\,m for bike, 5.00\,m for dumpster at 20\,m) reflect the scale ambiguity of monocular depth without metric grounding.

\begin{table}[!htb]
\centering
\vspace{0.5em}
\caption{Closed-loop performance on the Husky\,A200 ($N=8$/scenario; mean\,$\pm$\,std). SR: success rate; GHE: goal heading error; GDE: goal distance error at lock; Fr\'echet vs.\ teleoperation; BFA: behavior following accuracy.
}
\label{tab:closed_loop}

\scriptsize
\setlength{\tabcolsep}{2.9pt}
\renewcommand{\arraystretch}{1.25}

\begin{tabular}{@{}lcccccc@{}}
\toprule
& \textit{Scn.1} & \textit{Scn.2}
& \multicolumn{2}{c}{\textit{Scn.3}}
& \multicolumn{2}{c}{\textit{Scn.4}} \\
\cmidrule(lr){2-2}\cmidrule(lr){3-3}\cmidrule(lr){4-5}\cmidrule(lr){6-7}
Metric & Ours & Ours & Ours & BehAV & Ours & BehAV \\
\midrule

SR (\%)\,$\uparrow$
& 100
& 87
& \textbf{100} & 25
& \textbf{100} & 38 \\

GHE (rad)\,$\downarrow$
& 0.45\,$\pm$\,0.06
& 0.32\,$\pm$\,0.03
& \textbf{0.36\,$\pm$\,0.15} & 0.48
& \textbf{0.34\,$\pm$\,0.02} & 0.44 \\

GDE (m)\,$\downarrow$
& 0.04\,$\pm$\,0.02
& 0.01\,$\pm$\,0.01
& \textbf{0.01\,$\pm$\,0.01} & 3.38
& \textbf{0.01\,$\pm$\,0.01} & 1.50 \\

Fr\'echet (m)\,$\downarrow$
& 1.69\,$\pm$\,0.33
& 1.91\,$\pm$\,1.51
& \textbf{2.06\,$\pm$\,1.84} & 3.76
& \textbf{1.26\,$\pm$\,0.17} & 4.78 \\

BFA (\%)\,$\uparrow$
& ---
& 95.5\,$\pm$\,4.8
& 82.3\,$\pm$\,12.2 & \textbf{85.3}
& \textbf{91.9\,$\pm$\,2.4} & 90.7 \\
\bottomrule
\end{tabular}

\vspace{-2.8em}
\end{table}

\subsection{Full-System Closed-Loop Deployment}

Table~\ref{tab:closed_loop} reports closed-loop performance across the four scenarios. Edge-BehAV reaches the goal on 31/32 trials (96\% success rate). 
The goal distance error at first metric lock is 0.04\,m in \textit{Scn.1} and 0.01\,m in \textit{Scn.2--4} — matching, and at the stationary measurement pose slightly improving on, the 0.20\,m component MAE (Table~\ref{tab:distance}) across short (14\,m) and long (26\,m) range and the bearing-to-metric handover (\textit{Scn.2} and~\textit{Scn.4}). This is expected: the stop-and-measure state machine (Sec.~\ref{sec:method}) latches the goal from a fresh image and scan captured at rest, removing the motion-induced error of the moving-platform benchmark. Goal heading error stays within 0.32--0.45\,rad, and Behavior Following Accuracy is 95.5\%, 82.3\%, and 91.9\% for \textit{Scn.2--4}; the lower \textit{Scn.3} value and its spread come from a wide cone detour in a trial (BFA 50.7\%).

The \textit{Scn.3} and \textit{Scn.4} columns of Table~\ref{tab:closed_loop} place Edge-BehAV against the original BehAV system on the two disambiguation scenarios. This is a system-level, not strictly controlled, comparison: the released BehAV planner takes its goal from a one-shot operator-entered cloud distance estimate and lacks a LiDAR obstacle term, so it reads as evidence of end-to-end deployability rather than an isolated ablation — the controlled comparisons remain Tables~\ref{tab:slm_main}, \ref{tab:detection}, and~\ref{tab:distance}. Edge-BehAV succeeds on 31/32 trials, whereas BehAV succeeds on 25\% (\textit{Scn.3}) and 38\% (\textit{Scn.4}). The failure mode is the goal estimate: BehAV's GPT distance is off by 3.38\,m at 14\,m and 1.50\,m at 26\,m with large per-trial spread (1.06--6.08\,m across \textit{Scn.3}), so the robot stops well short of or past the goal. Obstacle avoidance is comparable (BFA within a few points), so the decisive advantage is metric goal grounding: Edge-BehAV re-grounds the object and updates range and bearing frame by frame, correcting transient errors before they accumulate.


\subsection{Discussion}

The results support the central claim that behavior-aware instruction decomposition can move off the cloud without sacrificing accuracy: the best offline model is statistically on par with the strongest cloud API (0.775 vs.\ 0.765, overlapping confidence intervals) at an order-of-magnitude lower latency, while the original BehAV GPT-4 backend reaches only 0.512. Residual difficulty concentrates in Contextual/Adversarial and adverse-terrain instructions requiring spatial disambiguation or uncommon vocabulary, a challenge that does not scale with model size in the 1--8B range and may be mitigated in deployment by more explicit operator phrasing, a burden we treat as a mitigation to be validated rather than assumed. The closed-loop trials show the three modules compose into reliable navigation on real hardware; because Edge-BehAV and BehAV avoid obstacles comparably, the success-rate gap is attributable almost entirely to sensor-grounded metric goal estimation.


\section{Conclusion}\label{sec:conclusion}
This paper presented Edge-BehAV, a fully onboard VLN system for outdoor mobile robots in connectivity-constrained environments. To enable fully onboard VLN, we systematically benchmarked edge-deployable SLMs for robotic navigation instruction decomposition, developed a hybrid semantic-geometric goal localization framework, and integrated these advances into a complete onboard navigation system.

Experimental results demonstrated cloud-level navigation instruction decomposition using onboard SLMs, reduced the mean goal localization error from 2.05 m to 0.20 m, and successful completion of 31 of 32 closed-loop outdoor navigation trials. Together, these results demonstrate that combining edge-deployable foundation models with sensor-grounded geometric reasoning provides a practical path toward fully onboard VLN for outdoor environments.

\balance
\bibliographystyle{IEEEtran}
\bibliography{bibliography}

@inproceedings{weerakoon2025behav,
  title={Behav: Behavioral rule guided autonomy using vlms for robot navigation in outdoor scenes},
  author={Weerakoon, Kasun and Elnoor, Mohamed and Seneviratne, Gershom and Rajagopal, Vignesh and Arul, Senthil Hariharan and Liang, Jing and Jaffar, Mohamed Khalid M and Manocha, Dinesh},
  booktitle={2025 IEEE International Conference on Robotics and Automation (ICRA)},
  pages={7044--7051},
  year={2025},
  organization={IEEE}
}

@inproceedings{8578485,
  author={Anderson, Peter and Wu, Qi and Teney, Damien and Bruce, Jake and Johnson, Mark and Sünderhauf, Niko and Reid, Ian and Gould, Stephen and van den Hengel, Anton},
  booktitle={2018 IEEE/CVF Conference on Computer Vision and Pattern Recognition},
  title={Vision-and-Language Navigation: Interpreting Visually-Grounded Navigation Instructions in Real Environments},
  year={2018},
  pages={3674-3683},
  doi={10.1109/CVPR.2018.00387}
}

@inproceedings{gu-etal-2022-vision,
    title = "Vision-and-Language Navigation: A Survey of Tasks, Methods, and Future Directions",
    author = "Gu, Jing  and
      Stefani, Eliana  and
      Wu, Qi  and
      Thomason, Jesse  and
      Wang, Xin",
    booktitle = "Proceedings of the 60th Annual Meeting of the Association for Computational Linguistics (Volume 1: Long Papers)",
    month = may,
    year = "2022",
    address = "Dublin, Ireland",
    publisher = "Association for Computational Linguistics",
    doi = "10.18653/v1/2022.acl-long.524",
    pages = "7606--7623",
}

@inproceedings{ahn2022saycan,
  title={Do as i can, not as i say: Grounding language in robotic affordances},
  author={Brohan, Anthony and Chebotar, Yevgen and Finn, Chelsea and Hausman, Karol and Herzog, Alexander and Ho, Daniel and Ibarz, Julian and Irpan, Alex and Jang, Eric and Julian, Ryan and others},
  booktitle={Conference on robot learning},
  pages={287--318},
  year={2023},
  organization={PMLR}
}

@inproceedings{shah2023lmnav,
  author    = {Shah, Dhruv and Osi{\'n}ski, B{\l}a{\.z}ej and ichter, brian and Levine, Sergey},
  title     = {{LM-Nav}: Robotic Navigation with Large Pre-Trained Models of Language, Vision, and Action},
  booktitle = {Conference on Robot Learning},
  pages     = {492--504},
  year      = {2023},
  publisher = {PMLR}
}

@inproceedings{shah2023vint,
  title={ViNT: A Foundation Model for Visual Navigation},
  author={Shah, Dhruv and Sridhar, Ajay and Dashora, Nitish and Stachowicz, Kyle and Black, Kevin and Hirose, Noriaki and Levine, Sergey},
  booktitle={Conference on Robot Learning},
  pages={711--733},
  year={2023},
  organization={PMLR}
}

@INPROCEEDINGS{10610665,
  author={Sridhar, Ajay and Shah, Dhruv and Glossop, Catherine and Levine, Sergey},
  booktitle={2024 IEEE International Conference on Robotics and Automation (ICRA)},
  title={NoMaD: Goal Masked Diffusion Policies for Navigation and Exploration},
  year={2024},
  pages={63-70},
  doi={10.1109/ICRA57147.2024.10610665}
}

@INPROCEEDINGS{10802716,
  author={Sathyamoorthy, Adarsh Jagan and Weerakoon, Kasun and Elnoor, Mohamed and Zore, Anuj and Ichter, Brian and Xia, Fei and Tan, Jie and Yu, Wenhao and Manocha, Dinesh},
  booktitle={2024 IEEE/RSJ International Conference on Intelligent Robots and Systems (IROS)},
  title={CoNVOI: Context-aware Navigation using Vision Language Models in Outdoor and Indoor Environments},
  year={2024},
  pages={13837-13844},
  doi={10.1109/IROS58592.2024.10802716}
}

@inproceedings{radford2021clip,
  author    = {Radford, Alec and Kim, Jong Wook and Hallacy, Chris and Ramesh, Aditya and Goh, Gabriel and Agarwal, Sandhini and Sastry, Girish and Askell, Amanda and Mishkin, Pamela and Clark, Jack and Krueger, Gretchen and Sutskever, Ilya},
  title     = {Learning Transferable Visual Models From Natural Language Supervision},
  booktitle = {Proceedings of the 38th International Conference on Machine Learning},
  pages     = {8748--8763},
  year      = {2021},
  publisher = {PMLR}
}

@inproceedings{luddecke2022clipseg,
  author    = {L\"{u}ddecke, Timo and Ecker, Alexander},
  title     = {Image Segmentation Using Text and Image Prompts},
  booktitle = {Proceedings of the IEEE/CVF Conference on Computer Vision and Pattern Recognition},
  pages     = {7086--7096},
  year      = {2022}
}

@inproceedings{xiao2024florence2,
  author    = {Xiao, Bin and Wu, Haiping and Xu, Weijian and Dai, Xiyang and Hu, Houdong and Lu, Yumao and Zeng, Michael and Liu, Ce and Yuan, Lu},
  title     = {Florence-2: Advancing a Unified Representation for a Variety of Vision Tasks},
  booktitle = {Proceedings of the IEEE/CVF Conference on Computer Vision and Pattern Recognition},
  year      = {2024}
}

@inproceedings{liu2023groundingdino,
  author    = {Liu, Shilong and Zeng, Zhaoyang and Ren, Tianhe and Li, Feng and Zhang, Hao and Yang, Jie and Li, Chunyuan and Yang, Jianwei and Su, Hang and Zhu, Jun and Zhang, Lei},
  title     = {Grounding {DINO}: Marrying {DINO} with Grounded Pre-Training for Open-Set Object Detection},
  booktitle = {Proceedings of the European Conference on Computer Vision},
  year      = {2024}
}

@article{zhang2023mobilesam,
  author    = {Zhang, Chaoning and Han, Dongshen and Qiao, Yu and Kim, Jung Uk and Bae, Sung-Ho and Lee, Seungkyu and Hong, Choong Seon},
  title     = {Faster Segment Anything: Towards Lightweight {SAM} for Mobile Applications},
  journal   = {arXiv preprint arXiv:2306.14289},
  year      = {2023}
}

@inproceedings{kirillov2023sam,
  author    = {Kirillov, Alexander and Mintun, Eric and Ravi, Nikhila and Mao, Hanzi and Rolland, Chloe and Gustafson, Laura and Xiao, Tete and Whitehead, Spencer and Berg, Alexander C. and Lo, Wan-Yen and Doll{\'{a}}r, Piotr and Girshick, Ross},
  title     = {Segment Anything},
  booktitle = {Proceedings of the IEEE/CVF International Conference on Computer Vision},
  pages     = {4015--4026},
  year      = {2023}
}

@misc{zhao2023fast,
      title={Fast Segment Anything},
      author={Xu Zhao and Wenchao Ding and Yongqi An and Yinglong Du and Tao Yu and Min Li and Ming Tang and Jinqiao Wang},
      year={2023},
      eprint={2306.12156},
      archivePrefix={arXiv},
      primaryClass={cs.CV}
}

@inproceedings{frantar2023gptq,
  author    = {Frantar, Elias and Ashkboos, Saleh and Hoefler, Torsten and Alistarh, Dan},
  title     = {{GPTQ}: Accurate Post-Training Quantization for Generative Pre-trained Transformers},
  booktitle = {Proceedings of the 11th International Conference on Learning Representations},
  year      = {2023}
}

@article{qwen2025report,
  author    = {{Qwen Team}},
  title     = {{Qwen2.5} Technical Report},
  journal   = {arXiv preprint arXiv:2412.15115},
  year      = {2025}
}

@article{google2025gemma3,
  author    = {{Google DeepMind}},
  title     = {Gemma 3 Technical Report},
  journal   = {arXiv preprint arXiv:2503.19786},
  year      = {2025}
}

@article{abdin2024phi3,
  author    = {Abdin, Marah and Jacobs, Sam Ade and Awan, Ammar Ahmad and Aneja, Jyoti and Awadallah, Ahmed and Awadalla, Hany and Bach, Nguyen and Bahree, Amit and Bakhtiari, Arash and others},
  title     = {Phi-3 Technical Report: A Highly Capable Language Model Locally on Your Phone},
  journal   = {arXiv preprint arXiv:2404.14219},
  year      = {2024}
}

@article{lu2024slmsurvey,
  author    = {Lu, Zhenyan and Li, Xiang and Cai, Dongqi and Yi, Rongjie and Liu, Fangming and Zhang, Xiwen and Lane, Nicholas D. and Xu, Mengwei},
  title     = {Small Language Models: Survey, Measurements, and Insights},
  journal   = {arXiv preprint arXiv:2409.15790},
  year      = {2024}
}

@article{chu2023mobilevlm,
  author    = {Chu, Xiangxiang and Qiao, Limeng and Lin, Xinyang and Xu, Shuang and Yang, Yang and Hu, Yiming and Wei, Fei and Zhang, Xinyu and Zhang, Bo and Wei, Xiaolin and Shen, Chunhua},
  title     = {{MobileVLM}: A Fast, Strong and Open Vision Language Assistant for Mobile Devices},
  journal   = {arXiv preprint arXiv:2312.16886},
  year      = {2023}
}

@inproceedings{vora2020pointpainting,
  author    = {Vora, Sourabh and Lang, Alex H. and Helou, Bassam and Beijbom, Oscar},
  title     = {{PointPainting}: Sequential Fusion for {3D} Object Detection},
  booktitle = {Proceedings of the IEEE/CVF Conference on Computer Vision and Pattern Recognition},
  pages     = {4604--4612},
  year      = {2020}
}

@inproceedings{qi2018frustum,
  author    = {Qi, Charles R. and Liu, Wei and Wu, Chenxia and Su, Hao and Guibas, Leonidas J.},
  title     = {Frustum {PointNets} for {3D} Object Detection from {RGB-D} Data},
  booktitle = {Proceedings of the IEEE/CVF Conference on Computer Vision and Pattern Recognition},
  pages     = {918--927},
  year      = {2018}
}

@inproceedings{peng2023openscene,
  author    = {Peng, Songyou and Genova, Kyle and Jiang, Chiyu and Tagliasacchi, Andrea and Pollefeys, Marc and Funkhouser, Thomas},
  title     = {{OpenScene}: {3D} Scene Understanding with Open Vocabularies},
  booktitle = {Proceedings of the IEEE/CVF Conference on Computer Vision and Pattern Recognition},
  year      = {2023}
}

@article{williams2018mppi,
  author    = {Williams, Grady and Drews, Paul and Goldfain, Brian and Rehg, James M. and Theodorou, Evangelos A.},
  title     = {Information-Theoretic Model Predictive Control: Theory and Applications to Autonomous Driving},
  journal   = {IEEE Transactions on Robotics},
  volume    = {34},
  number    = {6},
  pages     = {1603--1622},
  year      = {2018}
}

@INPROCEEDINGS{nuscenes,
  title={nuScenes: A multimodal dataset for autonomous driving},
  author={Holger Caesar and Varun Bankiti and Alex H. Lang and Sourabh Vora and 
          Venice Erin Liong and Qiang Xu and Anush Krishnan and Yu Pan and 
          Giancarlo Baldan and Oscar Beijbom}, 
  booktitle={CVPR},
  year=2020
}

\end{document}